\documentclass[journal]{IEEEtran}
\usepackage{booktabs}
\usepackage{caption}
\ifCLASSINFOpdf
\else
\fi
\usepackage{array}

\usepackage{amsfonts,amssymb}
\usepackage{graphicx}
\usepackage{color}
\usepackage{algorithmicx,algorithm}
\usepackage{multirow}
\begin{document}
\title{ETC: Temporal Boundary Expand then Clarify for Weakly Supervised Video Grounding with Multimodal Large Language Model}

\author{Guozhang~Li,
        Xinpeng~Ding,
        De~Cheng,
        Jie~Li,
        Nannan~Wang,~\IEEEmembership{Member,~IEEE},
        Xinbo~Gao,~\IEEEmembership{Fellow,~IEEE}
        }


\markboth{Journal of \LaTeX\ Class Files,~Vol.~14, No.~8, August~2015}%
{Shell \MakeLowercase{\textit{et al.}}: Bare Demo of IEEEtran.cls for IEEE Journals}



\maketitle

\begin{abstract}
Early weakly supervised video grounding (WSVG) methods often struggle with incomplete boundary detection due to the absence of temporal boundary annotations.
To bridge the gap between video-level and boundary-level annotation, explicit-supervision methods, \emph{i.e.} , generating pseudo-temporal boundaries for training, have achieved great success.
However, data augmentations in these methods might disrupt critical temporal information, yielding poor pseudo boundaries.
In this paper, we propose a new perspective that maintains the integrity of the original temporal content while introducing more valuable information for expanding the incomplete boundaries. 
To this end, we propose \textbf{EtC} (\textbf{E}xpand \textbf{t}hen \textbf{C}larify), first use the additional information to expand the initial incomplete pseudo boundaries, and subsequently refine these expanded ones to achieve precise boundaries.
Motivated by video continuity, \emph{i.e.} , visual similarity across adjacent frames, we use powerful multimodal large language models (MLLMs) to annotate each frame within initial pseudo boundaries, yielding more comprehensive descriptions for expanded boundaries.
% However, the MLLM-generated descriptions inevitably introduce some irrelevant information, resulting in overinclusive boundaries.
%
% Finally, we leverage a learnable approach to strike a balance between accurate yet incomplete boundaries and overly comprehensive but noisy ones, achieving more precise boundary delineation.
To further clarify the noise of expanded boundaries, we combine mutual learning with a tailored proposal-level contrastive objective to use a learnable approach to harmonize a balance between incomplete yet clean (initial) and comprehensive yet noisy (expanded) boundaries for more precise ones.
% integrate the information of both the initial and expanded bound
% strike a balance between accurate yet incomplete boundaries and overly comprehensive but noisy ones, achieving more precise boundary delineation.
% we 
% to integrate the information of both the initial and expanded boundaries and (ii)  to refine generated pseudo-boundaries in a learnable way.
%
Experiments demonstrate the superiority of our method on two challenging WSVG datasets.  
\end{abstract}


% \fi
\IEEEpeerreviewmaketitle

\section{Introduction}
\label{sec:intro}
Weakly Supervised Video Grounding (WSVG) aims to locate the moments of interest in a video based on a query description without temporal boundary annotations. 
Compared to its fully-supervised counterpart, weakly supervised approaches simplify the data collection process and avoid annotation biases,
finding vast applications in video action analysis \cite{li2023boosting}, video question and answer \cite{antol2015vqa,lei2018tvqa}, and human-computer interaction systems \cite{lei2019tvqa+,wang2021improving}.

WSVG methods typically fall into two categories: implicit supervision methods and explicit supervision methods. 
Implicit supervision methods, due to the absence of boundary annotations, approach moment localization as a form of proposal selection through proxy tasks~\cite{mithun2019weakly,lin2020weakly,zhang2020learning,ma2020vlanet,zheng2022weakly1}. 
These methods generally begin by generating proposals and then aiming to identify the proposal that most closely aligns with the given query by MIL-based~\cite{mithun2019weakly,huang2021cross,zhang2020counterfactual} or query-based reconstruction methods~\cite{lin2020weakly,ma2020vlanet,zheng2022weakly,zheng2022weakly1}.
\begin{figure}[!tbp]
\begin{center}
    \includegraphics[width=1.0\linewidth]{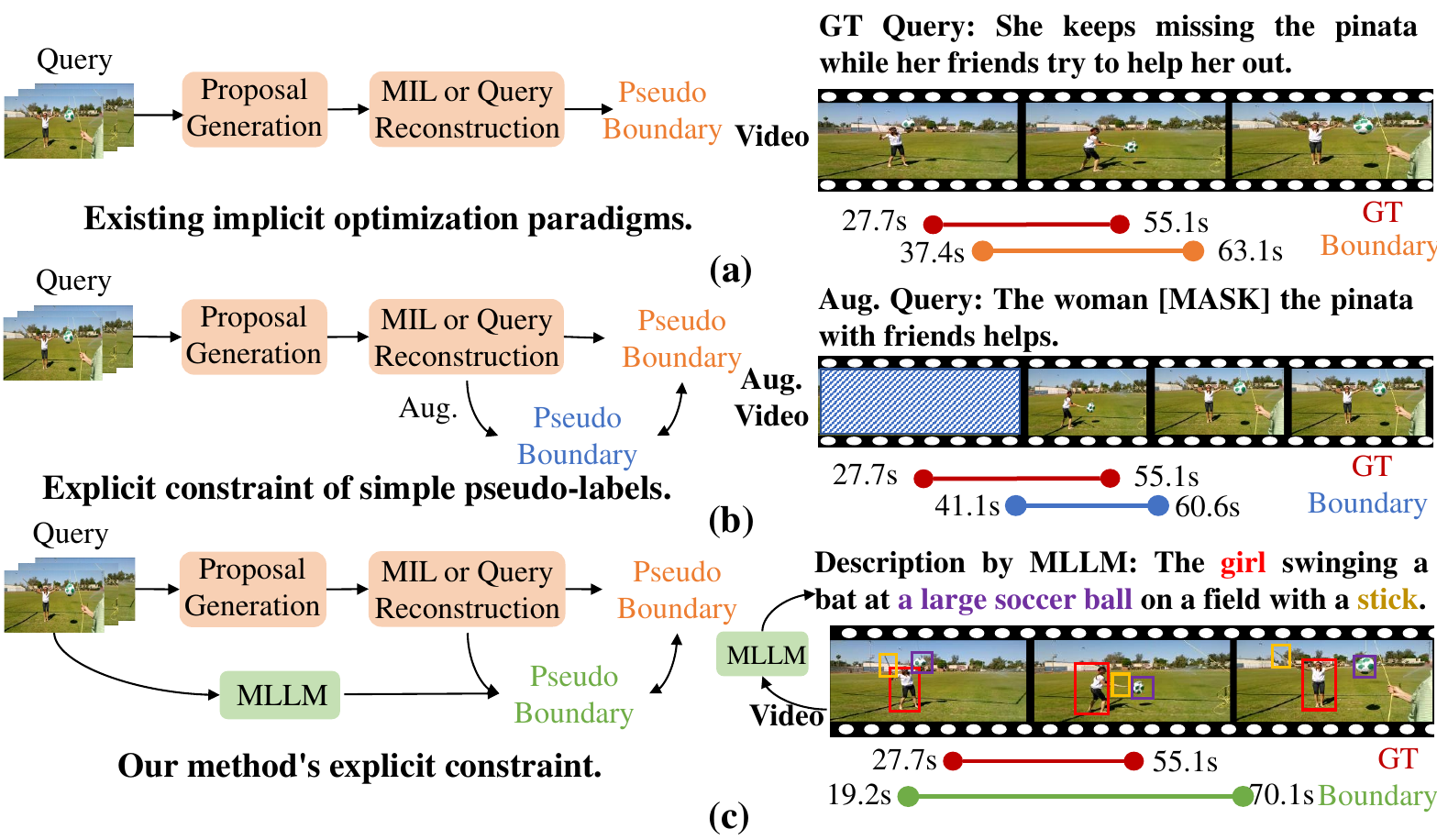}
    \vspace{-0.1cm}
\end{center}
   \caption{(a) The original implicit supervision methods (b) The original explicit supervision methods with simple pseudo label.  (c) The proposed method. The proposal result of description generated by the MLLM, where the proposal will be as complete as possible to cover all content.}
\label{fig1}
\vspace{-0.1cm}
\end{figure}
However, these methods tend to overly focus on the most distinct parts of the video or query, leading to sub-optimal WSVG models and issues with partial grounding.

Explicit supervision-based WSVG methods~\cite{huang2023weakly} have emerged to tackle existing limitations by generating pseudo-temporal boundaries, thus bridging the gap between video and boundary-level understanding. 
These methods commence by creating pseudo-labels through data augmentation and a mean teacher approach, utilizing these labels to refine the outputs of the basic WSVG model.
Data augmentation plays a crucial role in these methods, significantly influencing the quality of the pseudo-labels generated.
However, in video data augmentation, there's a risk of losing segments critical to the ground truth (Fig.\ref{fig1}(b)). Similarly, augmenting text might result in the omission of tokens essential for localization.
With the inherently sub-optimal WSVG model, the initial pseudo-labels are often imprecise and missing complete GT boundaries (e.g., the \textcolor{blue}{blue line}). 
Feeding augmented inputs that lose crucial information into the sub-optimal WSVG models may result in progressively poorer pseudo-labels, and deteriorating model performance.

We present a new perspective in this paper: to improve sub-optimized localization models, it's crucial to supply additional valuable information. This is in contrast to existing methods, which often risk losing critical information.
To this end, we introduce a two-step method, \textbf{EtC} (\textbf{E}xpand \textbf{t}hen \textbf{C}larify), which first expands the initial pseudo boundaries to the overinclusive boundaries with additional useful information and subsequently refines these expanded ones to achieve precise boundaries.
%
% To this end, we propose a two-step method named \textbf{EtC} (\textbf{E}xpand \textbf{t}hen \textbf{C}larify), to first expand the original ground truth (GT) with more relevant information and then refine these expanded segments, removing noise to achieve more precise boundaries.

%
Multimodal Large Language Models (MLLMs)~\cite{liu2023visual,chen2023videollm,hong20233d} have recently shown powerful zero-shot ability in generating more nuanced and detailed descriptions, surpassing conventional vision-language models~\cite{li2019visualbert,hu2021unit,wang2022object}.
Inspired by this, we explore leveraging MLLMs to obtain meaningful information to expand the initial pseudo boundaries.
Despite the initial pseudo-boundaries being incomplete, they are indeed located near significant frames within the GT boundaries; see Fig. \ref{fig:GT}. Given the continuity in videos, where visual elements in adjacent frames are typically similar, the frames within these pseudo-boundaries generally contain most of the visual information present in the GT boundaries. Based on this, we utilize MLLMs to thoroughly annotate each frame within the pseudo-boundaries, thereby obtaining descriptions that encompass a broader range of visual elements compared to the original GT query, as shown in Fig.\ref{fig1}(c).
Then, we use the MLLM-generated descriptions as input to the basic WSVG model, obtaining an expanded boundary, compared with the initial one (\textcolor{green}{green line} vs \textcolor{blue}{blue line} in Fig.~\ref{fig1}).
%
% Although initial pseudo boundaries are incomplete, they can actuall
% The next challenge resides in how to obtain meaningful information to expand the initial pseudo boundaries.

% we first collect the frames within the initial pseudo boundaries. 

% we first leverage MLLMs to generate descriptions for each frame within the initial pseudo boundaries.
%

% we aim to explore MLLMs to generate additional informative information to obtain more complete boundaries.
%
% To this end, we introduce a boundary-expanding module (BoundExpan) to expand the initial pseudo boundaries generated by the basic sub-optimized model.
%
% Considering the video information is hard to be expanded, since we do not obtain the precise temporal boundaries for GT.
% Hence, we focus on obtaining more complete boundaries by expanding the text query information with MLLMs.
%
% Specifically, given the initial pseudo boundaries, we leverage MLLMs to generate descriptions for each frame within the pseudo boundaries.
%
% As shown in Fig.\ref{fig1}(c), compared with GT queries, the MLLM-generated descriptions typically provide more comprehensive insights into people, objects, environments, and human interactions depicted in video frames.
%

Although the descriptions generated by MLLMs can enhance the GT query with more useful information, this process inevitably introduces some irrelevant information, leading to overinclusive boundaries. 
To clean the noisy boundaries, we propose the boundary clarify module equipped with two objectives, \emph{e.g.} , (i) a mutual learning objective to jointly consider the initial pseudo boundaries (incomplete yet clean) and the expanded pseudo boundaries (comprehensive yet noisy) and (ii) a novel proposal-level contrastive (PCL) object to use a learnable approach to harmonize a balance between those two kinds of boundaries for more precise ones.
% integrate the information of both the initial and expanded boundaries; (ii) a novel proposal-level contrastive (PCL) object to refine generated pseudo-boundaries in a learnable way by increases the similarity scores within the boundaries and decreases them outside.
%
% Our designed PCL utilizes both unimodal and multimodal alignment for obtaining the similarity scores, via LLMs (~\eg, Bert~\cite{devlin2018bert}) and MLLMs (~\eg,BLIP2~\cite{li2023blip}).
% To refine existing pseudo-boundaries, we employ a learnable approach that increases the similarity scores within the boundaries and decreases them outside. 
% For PCL, we generate a frame-level similarity sequence, where each frame's similarity score reflects its match degree with the GT. To enhance the accuracy of these scores, we utilize both multimodal and unimodal methods, allowing MLLMs and LLMs to jointly assess and score the similarity of each frame with the GT query.
%

In summary, this paper presents four key contributions: 
\begin{itemize}
    \item We introduce a new insight for WSVG \emph{i.e.}, enhancing sub-optimized models by supplying additional valuable information from MLLMs. To the best of our knowledge, it is the first to leverage MLLMs for video temporal localization.
    \item  We propose \textbf{EtC} (\textbf{E}xpand \textbf{t}hen \textbf{C}larify) which first expands the initial pseudo boundaries with additional information and subsequently refines these noisy expanded ones to achieve precise boundaries.
    \item  We design the novel proposal-level contrastive (PCL) object to clarify expanded pseudo-boundaries via both multimodal and unimodal alignment.
    \item  Extensive experiments on both the Charades-STA and ActivityNet Captions datasets demonstrate the efficacy of the proposed method in significantly improving the state-of-the-art performance of WSVG.
\end{itemize}

\section{Related Work}
\label{sec:rw}
\textbf{Fully-Supervised Video Grounding.} Many prior research employs supervised methods \cite{anne2017localizing,ge2019mac,liu2021context,paul2022text,yang2022video}.
To achieve precise moment localization via language description, it is essential for a video grounding model to implement cross-modal alignment of videos and sentences.
Current video grounding methods can be roughly divided into two categories.
The first is proposal-based pipeline \cite{gao2017tall,zhang2020learning,zhang2021multi,chen2019semantic,xiao2021boundary,ding2022exploring,wang2022negative}, which design pre-defined multi-scale proposals at the initial stage.
Then, after aggregating contextual multi-modal information, it predicts the alignment score of each proposal and finds the best matching moment.
In contrast, proposal-free pipeline \cite{mun2020local,chen2020rethinking,nan2021interventional,wang2020temporally,ding2021support,tang2021frame} are developed to regress the start and end timestamps directly.
These methods use multi-modal information to estimate the probability of each frame as a boundary frame, avoiding the time-consuming proposal ranking process.
Recently, some methods also propose diffusion-based generation methods, such as \cite{liang2023exploring,li2023momentdiff}.
However, the frame-level annotation is extremely laborious, this manual annotation process greatly hinders scalability and real-world applicability.

\label{sec:method}
\begin{figure*}[!tbp]
\begin{center}
    \includegraphics[width=1.0\linewidth]{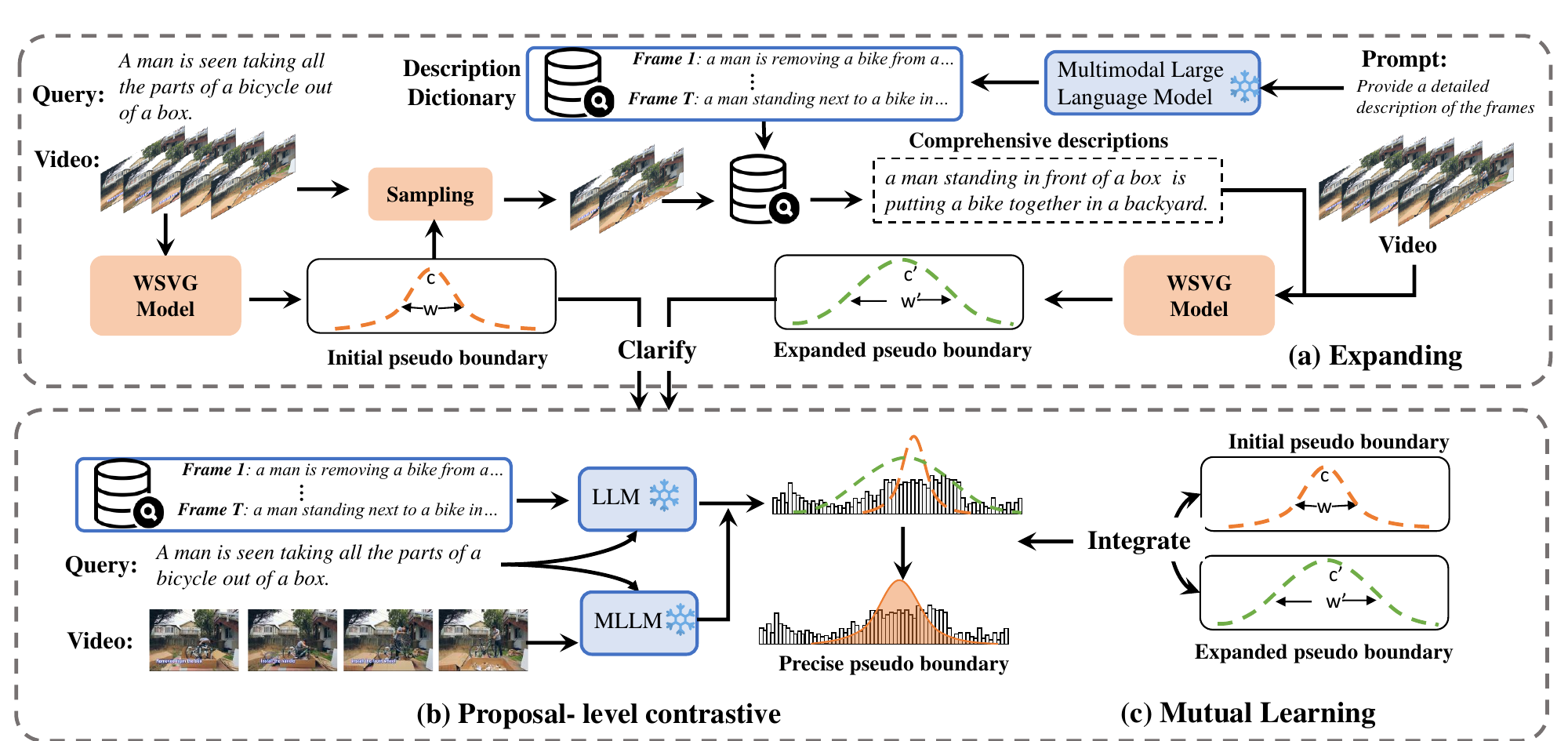}
    \vspace{-0.1cm}
\end{center}
   \caption{(a) The temporal pseudo boundary expand module. We utilize MLLM to describe frames within the initial pseudo boundary of the basic WSVG model to refine these expanded ones to guide the basic model to generate a comprehensive one. Then we clarify the expanded pseudo boundaries by (b) the PCL loss which harmonizes a balance between the initial pseudo boundary and expanded boundary and (c) the mutual learning jointly considers the initial pseudo boundaries (incomplete yet clean) and the expanded pseudo boundaries (comprehensive yet noisy.} %\xp{revise}.}
\label{fig2}
\vspace{-0.1cm}
\end{figure*}

\noindent \textbf{Weakly-Supervised Video Grounding.} 
Current WSVG methods generally generate potential proposals firstly and then employ query reconstruction or MIL based paradigm.
Specifically, query reconstruction based methods \cite{lin2020weakly,ma2020vlanet,song2020weakly,zheng2022weakly,zheng2022weakly1} select proposals based on cross-modal reconstruction loss, assuming text-matching proposals should best reconstruct the query.
Meanwhile, MIL-based methods \cite{gao2019wslln,mithun2019weakly,huang2021cross,zhang2020counterfactual,chen2022explore} learn the video-level cross-modal alignment by maximizing video-description match scores while suppressing unmatched descriptions.
However, these approaches lack explicit region-level constraints for proposals, leaving the model with the problem of partial grounding.
A recent work UGS \cite{huang2023weakly} designs a pseudo-label method based on mean-teacher framework with bayesian uncertainty, but it relies heavily on random data augmentation, resulting in unreliable pseudo-labels.
Unlike these methods, our method considers cyclical consistency between regions and text. We employ MLLMs to generate accurate descriptions for frames in the original grounding solution. 
These descriptions prompt the grounding model to produce comprehensive proposals, ensuring thorough coverage of event content and enhancing result accuracy.
Additionally, our proposed PLC loss refines proposals using the query-frame and query-description similarity obtained from MLLMs, effectively reducing the risk of error accumulation.
%Different from these methods, we utilize MLLM generates accurate and detailed descriptions for frames within the original grounding proposal,
%
%while the detailed description prompts the grounding model to generate proposals as complete as possible, ensuring comprehensive coverage of the mentioned content.
%
%We merge the complementary two proposals using mutual learning and employ a proposal-level contrastive loss to address the issue of error accumulation due to low-quality initial proposals.

% \vspace{-0.1cm}
\noindent \textbf{Visual-Language Pre-trained Model.}
Large scale visual-language pre-trained models such as CLIP \cite{radford2021learning}, ALIGN \cite{jia2021scaling} have demonstrated demonstrated strong general capabilities in multi-modal tasks \cite{huang2023clip2point,zou2023unim,ding2024holistic,ding2023hilm}.
Recently, enabling Large Language Models (LLM) (\emph{e.g.} GPT \cite{floridi2020gpt}, LLaMA \cite{touvron2023llama}) to understand multi-modal content has been increasingly studied.
Flamingo \cite{alayrac2022flamingo} inserts new cross-attention layers into the LLM to inject visual features, and pre-trains the new layers on billions of image-text pairs.
BLIP-2 \cite{li2023blip} effectively and efficiently leverage both frozen visual encoders and frozen LLMs for various vision-language tasks via Q-Former.
LLaVA \cite{liu2023visual} connected LLaMA with a visual encoder through a projection module, endowing it the ability to understand images.
In this paper, we focus on obtaining more complete boundaries by expanding the
text query information with MLLMs. 
and clarity the temporal boundaries by mutual learning strategy and proposal-level contrastive loss.

\newcommand{\ETC}{Expand then Clarify}
\newcommand{\IB}{Information Expand}
\newcommand{\BC}{Boundary Expand and Consistency}
\newcommand{\BR}{Boundary Clarify}
\section{The Proposed Method}
\subsection{overview}
The schematic illustration of our EtC framework is shown in Figure \ref{fig2}. 
We first employ MLLM to understand and describe the content of video frames guided by predefined prompts, and store these generated descriptions in a dedicated dictionary in the \IB   stage.
In our \BC  module, an input video-query pair is fed into a current WSVG model (\emph{e.g.} CPL \cite{zheng2022weakly1}), which produces initial pseudo temporal boundaries (grounding proposal).
We then retrieve detailed MLLM-generated descriptions for the frames within these pseudo boundaries from the description dictionary, which guides the WSVG model to generate expanded boundaries to cover all content of these description.
Subsequently, we clarity these two complementary temporal boundaries by through mutual learning to clean the noisy boundaries.
Meanwhile, we also clarify these temporal boundaries by the PCL. 
We generate a frame-level query-description and query-frame matching score via LLM and MLLM, and ensuring that the matching scores within the pseudo boundaries are higher than those outside.
our PCL further refines the pseudo boundaries generated by WSVG, 
thereby preventing the accumulation of errors caused by initial pseudo boundaries inaccuracies.

\subsection{Preliminary}
\noindent \textbf{Problem formulation.} 
For WSVG task, we are provided with a set of $N$ untrimmed videos defined as $\{{V_i}\}_{i=1}^N$, and their corresponding query sentences $\{{Q_i}\}_{i=1}^N$, each describing a moment of interest within the video.
The goal of WSVG model is to ground query each sentence to a specific temporal region in the corresponding $i$-th video with start timestamps $sta_i$ and end timestamps $end_i$.

\noindent \textbf{Feature Extraction.}
Given an untrimmed video $V$, we first feed them into a video encoder to obtain the embedded features $\mathbf{V} = \{\mathbf{v}_i\}_{i=1}^{T} \in \mathbb{R}^{T\times C}$, where $T$ is the number of sampled frames and $C$ is the feature dimension.
For the corresponding query sentence $Q$ of video $V$, the query feature is encoded as $\mathbf{Q} = \{\mathbf{q}_i\}_{i=1}^L \in \mathbb{R}^{L \times C}$, where $L$ is the number of words of the query.

\subsubsection{Text Information Expand}
\label{sec:3.3.2} 
We employ MLLM to generate auxiliary text information to describe each video frame, as shown in Figure \ref{fig2} (a).
The MLLM-generated descriptions typically provide more comprehensive insights into the frame content,
then these descriptions can help the WSVG model obtain a more general boundary.
Specifically, we used the open-source BLIP-2 \cite{li2023blip} pre-trained generative model. 
We use the following text prompts to generate multiple descriptions for all T video frames of the input video: 
(i) ``\emph{Generate captions for that video frame.} '' 
(ii) ``\emph{Provide a detailed description of the following frame.}'' 
(iii) ``\emph{Describe the following frame in detail.}'' 
(iv) ``\emph{Elaborate on the details of this frame in your own words.}'' 
(v) ``\emph{Describe the image concisely.}''
These prompts offer comprehensive descriptions of video frames. 
We generate 5 descriptions for each frame with the above prompt and store them as a dictionary before training. 
During training, we read the descriptions from the description dictionary. 
Then, we extract multiple descriptions of frames within the pseudo boundaries output by the basic WSVG model.
We then randomly sample one description to serve as the description for that pseudo boundaries region.

\subsection{Basic Grounding Model}
\label{sec:3.3.1}
We employ implicit supervision WSVG methods with learnable pseudo boundaries generation method \cite{zheng2022weakly,zheng2022weakly1} as our basic grounding model in our framework.

\noindent \textbf{Pseudo Temporal Boundaries Generation.} The most common method for pseudo boundaries generation in WSVG involves manually designing multi-scale sliding windows \cite{lin2020weakly,song2020weakly,zhang2020learning}. 
While simple in design, this approach is computationally inefficient and yields low-quality pseudo boundaries.
We follow \cite{zheng2022weakly,zheng2022weakly1} to conduct learnable Gaussian function pseudo boundaries.
Specifically, we append an additional [CLS] token at the end of video features, denoted as: $\hat{\mathbf{V}}=\{\mathbf{v}_1,...,\mathbf{v}_T,\mathbf{v}_{cls}\}$.
We utilize a transformer to fuse video and query features for cross-modal interaction, acquiring hidden layer feature $\hat{\mathbf{H}}=\{\mathbf{h}_1,...,\mathbf{h}_T,\mathbf{h}_{cls}\}$.
The output of the [CLS] token, $\mathbf{h}_{cls}$, is used to generate $N_p$ pseudo boundaries by employing a fully connected layer (FC) followed by the sigmoid function:
\begin{equation} 
\label{E1} 
{
\mathbf{p} = \{\mathbf{c},\mathbf{w}\} = Sigmoid(FC(\mathbf{h}_{cls}) \in \mathbb{R}^{N_p \times 2}
},
\end{equation}
where $\mathbf{c},\mathbf{w}$ denotes the center and width of the pseudo boundaries, respectively. 
Finally, the pseudo boundary feature $\mathbf{v}_p$ is generated by pooling video feature $\mathbf{V}$ with the temproal boundary $\mathbf{p}$.

\noindent \textbf{Query-Reconstruction Based Grounding Model.}  These methods \cite{lin2020weakly,ma2020vlanet,song2020weakly,zheng2022weakly} assume that proposals matching the text should best reconstruct the entire query. 
Typically, the original query sentence $Q = \{q_1,...,q_L\}$ has one-third of its total word length $L$ randomly masked with a specific `[MASK]' symbol to generate the masked sentence $\hat{Q}$.
A reconstructor then reconstructs the original query by predicting the current word conditionally using the proposal feature $\mathbf{v}p$ and the previously generated word features $\mathbf{\hat{Q}}{[0:l-1]}$.  
The final reconstruction loss is formulated as follows: 
\vspace{-0.1cm}
\begin{equation}
\label{E2}
\mathcal{L}_{g} = -\sum^{L}_{l=1} \log\mathbf{P}(q_i|\mathbf{v}_p,\mathbf{\hat{Q}}_{[0:l-1]})).
\end{equation}

\noindent \textbf{MIL-Based Grounding Model.}
These methods \cite{mithun2019weakly,zhang2020counterfactual,chen2022explore} optimize the grounding model by maximizing the difference in matching scores between positive video-query pairs and negative video-query pairs.
Specifically, for a matching video-query pair, a set of $N$ proposals is generated firstly.
We calculate the matching scores between the query feature $\mathbf{Q}$ and the video feature within proposals $\mathbf{v}_p$.
Then, the average matching scores of the top-k proposals are calculated as the video-level cross-modal matching score $M_{pos}$ for this video-text pair.
Similarly, the matching score $M_{neg}$ of a negative video-query pair can be obtained, the model loss function is formulated as:
\vspace{-0.1cm}
\begin{equation}
\label{E3}
\mathcal{L}_{g} = max(\Delta, M_{neg}-M_{pos}),
\end{equation}
\vspace{-0.1cm}
where $\Delta$ is the threshold.

\subsection{Boundary Expand and Classify framework}
\subsubsection{Temporal Boundary Expand}
Comparing with current explicit supervision WSVG methods based on data augmentation and mean teacher, we propose to supply additional information to improve WSVG model which reduce the risk of losing critical information.
In this paper, we explore leverage MLLMs to provide nuanced and detailed description for video frames to help expand initial grounding boundary.
Despite the absence of ground truth (GT) temporal boundaries in a weakly supervised setting, the excellent performance of existing WSVG models has positioned the generated pseudo-temporal boundaries near the effective frames within the GT boundaries.
Due to the continuity of video frames, frames within these pseudo boundaries also contain the majority of information within the GT boundaries. 
Therefore, we use MLLM to annotate frames within the initial pseudo boundaries, thereby obtaining a more detailed and rich description of elements within the video event.
Subsequently, we use the descriptions generated by MLLM as input for the base WSVG model, obtaining expanded boundaries.
Concretely, as shown in Figure \ref{fig2} (b), in our EtC method, there are three steps in the temporal boundaries expansion process:
(1) Generating initial pseudo boundaries $\mathbf{p}_o$ from a basic WSVG model according to the given video-query pair.
(2) Generating detailed descriptions for frames within the initial pseudo boundaries in (1). Here we retrieve the description corresponding to the frame in the dictionary in \ref{sec:3.3.2}.
(3) Generating new pseudo boundaries $\mathbf{p}_n$ using the basic grounding model of the same structure according to the video feature and the description generated in (2).
\begin{table*}[!tbp]
    \centering
    \small
    \caption{Comparisons with state-of-the-art methods on C-STA dataset and ANC dataset. `rep.' denotes our replicated result.}
    \vspace{-0.1cm}

    \begin{center}
    \scalebox{1.0}[1.0]{
    \begin{tabular}{c|c|c|ccc|ccc}
    \hline
     \multirow{2}{*}{Type}&\multirow{2}{*}{Method}&\multirow{2}{*}{Year} &\multicolumn{3}{c|}{Charades-STA} &\multicolumn{3}{c}{ActivityNet Captions} \\ 
     \cline{4-9}
    ~&~&~&R1@0.3&R1@0.5&R1@0.7&R1@0.1&R1@0.3&R1@0.5\\
     \hline 
    \multirow{3}{*}{Full} & DRN \cite{zeng2020dense} & CVPR20 & - & 53.09 & 31.75 & - & 45.45 & 24.36\\
    ~ & BPNet \cite{xiao2021boundary} &AAAI22 & 65.48 & 50.75 & 31.64 & - & 58.98 & 42.07 \\
    ~ & SCDM \cite{yuan2019semantic} & TPAMI22 & - & 54.33 & 33.43 & - & 54.80 & 36.75\\
     \hline 
    \multirow{15}{*}{Weak} &TGA \cite{mithun2019weakly}&CVPR19& 32.14 & 19.94 & 8.84  & - & - & -\\
    ~&CTF \cite{chen2020look}&arxiv20& 39.80 & 27.30 & 12.90 & 74.20 & 44.30 & 23.60\\
    ~&SCN \cite{lin2020weakly}&AAAI20&42.96 & 23.58 &9.97 & 71.48 & 47.23 & 29.22\\
    ~&BAR \cite{wu2020reinforcement}& MM20& 44.97 & 27.04 & 12.23 & - & 49.03 & 30.73\\
    ~&EC-SL \cite{chen2021towards}&CVPR20&-&-&-&68.48 &44.29&24.16\\
    ~&VLA \cite{ma2020vlanet}&ECCV20&45.24 & 31.83 & 14.17&- &-&- \\
    ~&CCL \cite{zhang2020counterfactual}& NIPS21 & - & 33.21 & 15.68 & -&50.12 &31.07 \\
    ~&CRM \cite{huang2021cross}& ICCV21 & 53.66 & 34.76 & 16.37 &81.61&55.26&32.19\\
    ~&VCA \cite{wang2021visual}&  MM21 & 58.58 & 38.13 & 19.57 &67.96&50.45&31.00\\
    ~&LCNet \cite{yang2021local}& TIP21 & 59.60 &39.19 & 18.87 &78.58 &48.49 & 26.33\\
    ~&CNM \cite{zheng2022weakly}& AAAI21 & 60.39 &35.43 & 15.45&78.13&55.68&33.33 \\
    ~&CPL \cite{zheng2022weakly1}& CVPR22 & 66.40 & 49.24 & 22.39&79.86 & 53.67 & 31.24 \\
    ~&UGS \cite{huang2023weakly}& CVPR23 & 69.16& 52.18 & 23.94&82.10 & 58.07 & 36.91\\
    ~&CCR \cite{lv2023counterfactual}&  MM23 & 68.59& 50.79 & 23.75 & 80.32 & 51.19 & 28.19\\
    ~&SCA \cite{yoon2023scanet}& ICCV23 & 68.04 & 50.85& 24.07 &83.62 &56.07 &31.52\\
     \hline
    Weak&CPL (rep.)&-&67.57&49.37&22.26 & 81.34 &51.18 &31.23\\
      Weak&CPL + Ours&-&\textbf{69.84} & \textbf{53.39} & \textbf{25.84} & \textbf{83.93} & \textbf{61.31} & \textbf{37.01}\\
     \hline     
    \end{tabular}
    }
    \end{center}
    \label{tab1}
\vspace{-0.1cm}
\end{table*}

\begin{table*}[!tbp]
    \centering
    \small
    \setlength{\abovecaptionskip}{-0.1cm}
    \caption{Ablation studies on C-STA and ANC datasets. `Base' denotes the baseline model CPL. `$\mathcal{L}_{m}$' denotes the mutual learning strategy, and `PCL' denotes the Proposal-based contrastive loss. `R1@$n$' denotes the metric `Rank-1@IOU $n$'.}
    \vspace{-0.1cm}
    \begin{center}
    \scalebox{1.0}[1.0]{
    \begin{tabular}{c|ccc|ccc|ccc}
    \toprule
     \multirow{2}{*}{Exp}&\multicolumn{3}{c|}{Components} &\multicolumn{3}{c|}{Charades-STA} &\multicolumn{3}{c}{ActivityNet Captions}
      \\ 
     \cline{2-10}
     ~&Base &$\mathcal{L}_{m}$ & PCL & R1@0.3 & R1@0.5 &R1@0.7& R1@0.1&R1@0.3 & R1@0.5 \\
        \hline
     1&$\checkmark$ & $\times$ & $\times$ & 66.40 & 49.24 & 22.39 &80.34 & 56.18 & 32.23\\
     2&$\checkmark$ & $\checkmark$ & $\times$ & 69.56 & 53.07 & 25.30 & 83.07 & 60.87 & 35.05\\
     3&$\checkmark$ & $\times$ & $\checkmark$ & 67.07 & 50.73 & 23.48 & 81.03 & 57.70 & 34.16\\
     4&$\checkmark$ & $\checkmark$ & $\checkmark$ & \textbf{69.84} & \textbf{53.39} & \textbf{25.84}  & \textbf{83.93} & \textbf{61.31} & \textbf{37.01}\\
     \bottomrule   
    \end{tabular}
    }
    \end{center}
    \label{tab3}
\vspace{-0.1cm}
\end{table*}

\subsubsection{Temporal Boundary Clarify.}
\label{sec:3.4}
\textbf{Mutual learning strategy.} Although the MLLM-generated descriptions provide more useful information, this process inevitably introduce some irrelevant information,leading to overcompleted boundaries.
This paper we propose to clean the boundaries with two objectives.
The first is mutual learning strategy, which jointly consider the initial incomplete pseudo boundaries and expanded over-complete pseudo boundaries.
By integrating two complementary pseudo temporal boundaries, we can achieve a precise grounding result.
Specifically, a consistency constraint loss is imposed between the two pseudo boundaries to learn from each other:
\begin{equation}
\label{E4}
\mathcal{L}_{m} = MSE(\mathbf{p}_o,\psi(\mathbf{p}_n) + MSE(\mathbf{p}_n,\psi(\mathbf{p}_o),
\end{equation}
where, `MSE$(\cdot,\cdot)$' denotes the Mean Squared Error loss and  $\psi(\cdot)$ represents a truncated input gradient function.

\noindent \textbf{Proposal-level contrastive loss.} We further clarify the obtained pseudo boundaries by the tailored PCL, as shown in Figure \ref{fig2} (c).
The PCL aims to clarify the expanded pseudo label via both multi-modal and uni-modal alignment.
Specifically, we calculate the query-frame matching (QFM) score and query-description matching (QDM) score via MLLM and LLM respectively.
As shown in Fig. \ref{fig:qcm}, video frames within a temporal boundary that is usually the most suitable for a query have high QFM and QDM scores.
The accuracy of the extended pseudo boundaries relies heavily on the detailed description generated based on the initial pseudo boundaries area video frame.
Hence, we impose the PCL to improve the quality of the initial pseudo-boundary, thus provide the more precise and detailed annotation of the GT's events, which improves the quality of the extended pseudo-boundary.
Specifically, for a given video-query pair, we first read the descriptions of each frame within the video from the description dictionary in \ref{sec:3.3.2}.
Then we calculate the similarity of the original query to the descriptions of each frame as the query-description match score (QDM) $S$ for that frame by a LLM, such as Bert \cite{devlin2018bert}, 
and the video-level query-description match score sequence is expressed as $\mathbf{S} = \{s_i\}_{i=1}^T$.
Besides, the query-frame matching score (QFM) sequence $\mathbf{S}^f$ can be also calculated from the input query and each frame of input video by the MLLM.
The PCL loss aims to harmonize a balance between those two kinds of learnable boundaries $p_o=\{c,w\}$ for more precise ones. 
ensuring that the mean matching score $\bar{\mathbf{S}}_{in}$ and $\bar{\mathbf{S}}^f_{in}$ within the boundaries is higher than that outside $\bar{\mathbf{S}}_{out}$ and $\bar{\mathbf{S}}^f_{out}$.
Here we take the use of QDM score in comparative learning loss as an example:
\begin{equation}
\label{E5}
\bar{\mathbf{S}}_{in} = \sum_{sta}^{end} {s_i},  \bar{\mathbf{S}}_{out1} = \sum_{sta-\tau w}^{sta} {s_i}, \bar{\mathbf{S}}_{out2} = \sum_{end} ^{end+\tau w} {s_i},
\end{equation}
where $sta = c-w/2$, $end = c+ w/2$ are the start and end timestamp for a pseudo boundaries $p$,
$c$ and $w$ are the center and width of the learnable pseudo boundaries $p$,
and the $\tau$ is a hyper-parameter.
The PCL loss can be formulated as:
\begin{equation}
\label{E6}
\mathcal{L}_{c} = max(\bar{\mathbf{S}}_{ou1}-\bar{\mathbf{S}}_{in},\delta) +  max(\bar{\mathbf{S}}_{ou2}-\bar{\mathbf{S}}_{in},\delta),
\end{equation}
where $\delta$ is the threshold of the contrastive loss.
Similarly, we can also get the contrast loss $\mathcal{L}_{c}^{f}$ when using QFM scores.

\subsection{Model Training and Inference}
\noindent \textbf{Optimizing Process.} Considering all the aforementioned
objectives, our final objective function of the whole framework arrives at:
\begin{equation}
\label{E7}
\mathcal{L} = \mathcal{L}_{go} + \mathcal{L}_{gn} + \alpha \mathcal{L}_{m} + \beta (\mathcal{L}_{c} + \mathcal{L}_{c}^{f}),
\end{equation}
where $\mathcal{L}_{go}$ and $\mathcal{L}_{gn}$ are the loss functions of the two base grounding models, 
while $\mathcal{L}_{m}$, $\mathcal{L}_{c}$ are the mutual learning loss and proposal-level contrastive loss for the proposed framework,
$\alpha, \beta$ are the hyper-parameters to balance these loss terms. 

\noindent \textbf{Model Inference.} In the context of inference, the previously learned grounding model is used to generate temporal boundaries. The result of the final precise pseudo boundaries of the grounding model is used as the final output.

%\clearpage
%\setcounter{page}{1}
%\maketitlesupplementary

\section{Experiment}
\label{sec:exp}

\begin{table*}[!tbp]
    \centering
    \caption{(a) Integrating our EtC framework to existing method on ANC dataset. `rep.' denotes our reproduced result. (b) The impact of EtC using different pseudo-label generation methods on ANC dataset. Pseudo-label$^1$: iterative pseudo-label strategy; Pseudo-label$^2$: Mean-Teacher; Pseudo-label$^3$: UGS's pseudo-label method.}
 
	\begin{minipage}{0.45\linewidth}
		\centering
\scalebox{1.2}[1.2]{
        \begin{tabular}{c|ccc}
        \hline
        method&R1@0.1&R1@0.3&R1@0.5\\ \hline
        CNM (rep.) \cite{zheng2022weakly}&78.39
         & 55.25 & 33.05\\
         CNM + Ours&\textbf{84.46} & \textbf{61.04} & \textbf{34.20}\\
         \hline
         CPL (rep.) \cite{zheng2022weakly1}&80.34 & 56.18 & 32.23\\
         CPL + Ours&\textbf{83.93} & \textbf{61.31} & \textbf{37.01}\\
         \hline
         RSTPN (rep.) \cite{zhang2020regularized} &71.85
         & 47.08 & 28.25\\
         RSTPN + Ours&71.65 & \textbf{50.81} & \textbf{31.11}\\
        \hline
        \end{tabular}
        }
        \caption*{(a) }
        \label{tab:Scal}
	\end{minipage}
	%\qquad
	\hfill
	\begin{minipage}{0.45\linewidth}
		\centering
\scalebox{1.2}[1.2]{
        \begin{tabular}{c|ccc}
        \hline
         \multirow{2}{*}{method}&\multicolumn{3}{c}{Rank-1@IoU(\%)} \\ 
         \cline{2-4}
         ~&0.1&0.3&  0.5\\
         \hline
        Base & 79.86&53.67	&31.24\\
        Base + pseudo-label$^1$ & 78.53 &55.34 &31.70\\
        Base + pseudo-label$^2$	& 80.48&56.55	&33.80\\
        Base + pseudo-label$^3$	&81.36 &58.72   &34.31\\
        Base + Our's MLLM   &\textbf{83.93} &\textbf{61.31}   &\textbf{37.01}\\
        \hline  
        \end{tabular}
        }
        \caption*{(b) }
         \label{tab:comp}
	    \end{minipage}
     \vspace{-0.2cm}
\end{table*}

\subsection{Experimental Setting}
\noindent \textbf{Datasets.} We conducted experiments on two publicly available datasets \textbf{Charades-STA (C-STA)} \cite{gao2017tall} and \textbf{ActivityNet Captions (ANC)} \cite{krishna2017dense}, to validate the performance of our EtC framework.
\textbf{C-STA} dataset is builut upon Charades dataset \cite{sigurdsson2016hollywood} by Gao \emph{et al}, %
which contains 5338/1334 videos and 12,408/3720 video-query pairs for training/testing. We report our results on the test split of the Charades-STA dataset. 
\textbf{ANC} is built upon ActivityNet v1.3 dataset \cite{caba2015activitynet}, which contains 10,009/4917/5044 videos and 37,417/17,505/17,031 video-query pairs for training/validation/testing. 
We follow the previous works \cite{gao2017tall,zheng2022weakly}, and report our results on the val-2 split.

\noindent \textbf{Evaluation Metric.} For assessing the pseudo boundaries performance, we employ the ``Rank-1 @ IoU $n$'' metric with a rank-1 recall rate.
A pseudo boundaries is considered correct if its Intersection over Union (IoU) with the ground-truth pseudo boundaries exceeds the predefined threshold $n$.
In accordance with previous methods, we also set $n = \{0.3,0.5,0.7\}$ for Charades-STA dataset and $n = \{0.1,0.3,0.5\}$ for ActivityNet Caption dataset.

\subsection{Implementation Details}
For the video encoder in grounding model, we follow previous methods \cite{zheng2022weakly1,yoon2023scanet} to utilize I3D \cite{DBLP:conf/cvpr/CarreiraZ17} for C-STA and C3D \cite{tran2015learning} for ANC.  
The features are extracted by
downsampling each video at a rate of 8, and the maximum video segments is set as 200.
We continue to use Pre-trained GloVe word2vec \cite{pennington2014glove} as the text encoder in the grounding model as pervious methods, and the maximum query word length is 20, 
We use BLIP-2 \cite{li2023blip} to generate descriptions for video frames.
We use Adam optimizer with learning
rate set to 0.0004 for training both networks, the learning rate is decayed with an inverse square root scheduler.
For the hyper-parameter of the proposed framework, we set $\tau=0.25$, $\delta=0.15$ for our PCL loss, and for $\alpha,\beta$, we set them as $0.5, 0.1$ for C-STA and $0.25, 0.05$ for ANC.
In addition, to ensure the grounding model produces high-quality pseudo boundaries for our method, 
We first warm up the grounding model with 7 epochs for C-STA and 3 epoch for ANC. The total training epoch for these datasets is 30.
When applying on grounding models that generate multiple pseudo boundaries, we only use the top-1 pseudo boundaries in our method.

\begin{table*}[!tbp]
    \centering
    \caption{Ablation studies of mutual learning strategy on ANC dataset, where the baseline model is the CPL.}
    
	\begin{minipage}{0.3\linewidth}
		\centering
    \renewcommand\tabcolsep{3pt}
    \scalebox{1.2}[1.2]{
        \begin{tabular}{c|ccc}
        \hline
        $n_{p}$&R1@0.1&R1@0.3&R1@0.5\\ \hline
        1 & 83.35	&58.94	&32.51\\ 
        5 &	83.93  &61.31   &37.01\\
        10& 84.12	&61.65	&34.82\\
        \hline
        \end{tabular}
        }
        \caption*{(a) The impact of the number of generated descriptions $n_p$ per frame on results.}
        \label{tab:np}
	\end{minipage}
	%\qquad
	\hfill
	\begin{minipage}{0.3\linewidth}
		\centering
    \renewcommand\tabcolsep{3pt}
    \scalebox{1.2}[1.2]{
        \begin{tabular}{c|ccc}
        \hline
        $n_{f}$&R1@0.1&R1@0.3&R1@0.5\\ \hline
        1 & 83.64	&59.12	&33.58\\ 
        5 &	83.93   &61.31 & 37.01\\
        10& 83.91	&61.26	&35.55\\
        \hline
        \end{tabular}
        }
        \caption*{(b) The impact of the number of sampled frames $n_f$ on results in description generation.}
         \label{tab:nf}
	    \end{minipage}
        \hfill
	\begin{minipage}{0.35\linewidth}
		\centering
    \renewcommand\tabcolsep{3pt}
    \scalebox{1.2}[1.2]{
        \begin{tabular}{c|ccc}
        \hline
        model&R1@0.1&R1@0.3&R1@0.5\\ \hline
        BLIP	&83.87 &61.47	&34.75\\
        BLIP2	&83.93 &61.31 & 37.01\\
        LLaVa   &82.86 &60.02   &34.35\\
        \hline
        \end{tabular}
        }
        \caption*{(c) The impact of utilizing region descriptions generated by different MLLMs on grounding results.}
         \label{tab:mllm}
        \end{minipage}
        \vspace{-0.2cm}
\end{table*}
\subsection{Comprasion with the State-of-the-Art}
We compare the proposed method with state-of-the-art weakly-supervised video grounding methods in Tab. \ref{tab1}.
For C-STA datasets, the proposed framework evidently outperforms current state-of-the-art WSVG approaches, especially in high IoU experimental settings. 
Specifically, On three important criterion: R1@IoU 0.3, R1@IoU 0.5 and R1@IoU 0.7, we surpass the state-of-the-art method \cite{huang2023weakly} by 0.7\%, 1.2\% and 1.9\%, respectively.
For the larger ANC dataset, our method still obtains significant improvement over existing the state-of-the-art weakly-supervised methods, 
especially on the Rank-1@IoU 0.1 and Rank-1@IoU 0.3 metric, which are 1.8\% and 3.3\% improvement over current state-of-the-art methods \cite{huang2023weakly}.
Besides, for the two datasets, compared with the baseline model CPL \cite{zheng2022weakly1}, the proposed framework can obtain about 3.7\% and 4.5\% improvement on average,
and even achieve performance comparable to fully supervised methods.

\begin{table*}[!tbp]
    \centering
    \caption{Ablation studies of PCL on ANC dataset, where the CPL as the baseline.}
    \begin{minipage}{0.31\linewidth}

	\centering
     \renewcommand\tabcolsep{3pt}
     \scalebox{1.2}[1.2]{
        \begin{tabular}{c|ccc}
        \hline
        PCL&R1@0.1&R1@0.3&R1@0.5\\ \hline
        w/o & 83.07 & 60.87 & 35.05\\
        $\mathcal{L}_{c}$ & 83.52 & \textbf{61.95} & 35.45\\
        $\mathcal{L}_{c}^{f}$ &83.32 & 60.88 & 35.88\\
        Both &\textbf{83.93} & 61.31 & \textbf{37.01}\\
        \hline
        \end{tabular}
        }
        \caption*{(a) The influence of using QFM and QDM for PCL loss.}
         \label{tab:qfm}
        \end{minipage}
        \hfill
	\begin{minipage}{0.3\linewidth}
		\centering
      \renewcommand\tabcolsep{3pt}
      \scalebox{1.2}[1.2]{
        \begin{tabular}{c|ccc}
        \hline
        $\tau$&R1@0.1&R1@0.3&R1@0.5\\ \hline
        0.1 & 81.38&59.35 &34.12\\ 
        0.15 & 83.15 & 61.11 & 36.09\\ 
        0.2& 83.51 & 60.37 & 35.59 \\
        0.25 &\textbf{83.93} & 61.31 & \textbf{37.01}\\
        \hline
        \end{tabular}
        }
        \caption*{(b) The impact of parameter $\tau$ on the final grounding results.}
        \label{tab:t}
	\end{minipage}
	%\qquad
	\hfill
	\begin{minipage}{0.31\linewidth}
		\centering
      \renewcommand\tabcolsep{3pt}
      \scalebox{1.2}[1.2]{
        \begin{tabular}{c|ccc}
        \hline
        $\delta$&R1@0.1&R1@0.3&R1@0.5\\ \hline
        0.1 & 83.97	&60.28	&33.57\\ 
        0.15 &\textbf{83.93} & 61.31 & \textbf{37.01}\\
        0.2& 82.43	&60.11	&34.84\\
        0.25& 81.53 & 60.86 & 36.77\\
        \hline
        \end{tabular}
        }
        \caption*{(c) The impact of parameter $\delta$ on the final grounding results.}
         \label{tab:de}
    \end{minipage}

\end{table*}

\subsection{Ablation study}
\noindent \textbf{Effectiveness of Each Component}
We analyze the effectiveness of each component in our EtC framework here.
We conduct ablation studies on the C-STA and ANC datasets in Tab. \ref{tab3}.
Specifically, we implement four variants of the proposed method as follows: (1) `Base': We use the query-reconstruction based method CPL \cite{zheng2022weakly1} in Section \ref{sec:3.3.1} as the baseline model. (2) `Base + mutual learning'. (3) `Base + PCL'. (4) The final framework.
The results of line 2 and 4 of Tab. \ref{tab3}  show the efficacy of mutual learning, which can obtain about 3.3\% and 3.6\% improvement on average for C-STA and ANC datasets, respectively.

The mutual learning strategy provides comprehensive explicit supervision at the regional level by leveraging the strategy of mutual learning and the multi-modal understanding capability of MLLM.
By integrating two complementary pseudo boundaries, our approach effectively enhances model performance.
The results of line 3 and 4 of Tab. \ref{tab3} show the efficacy of the PCL, which can obtain about 0.7\% and 0.6\% improvement on average for C-STA and ANC dataset, respectively.
The proposed PCL loss helps to adjust the location and width of the learnable pseudo boundaries, ensuring the pseudo boundaries can cover the video frames that better matches the query to harmonize a balance between those two kinds of boundaries for more precise ones.

\begin{figure}[!htbp]
\begin{center}
    \includegraphics[width=1.0\linewidth]{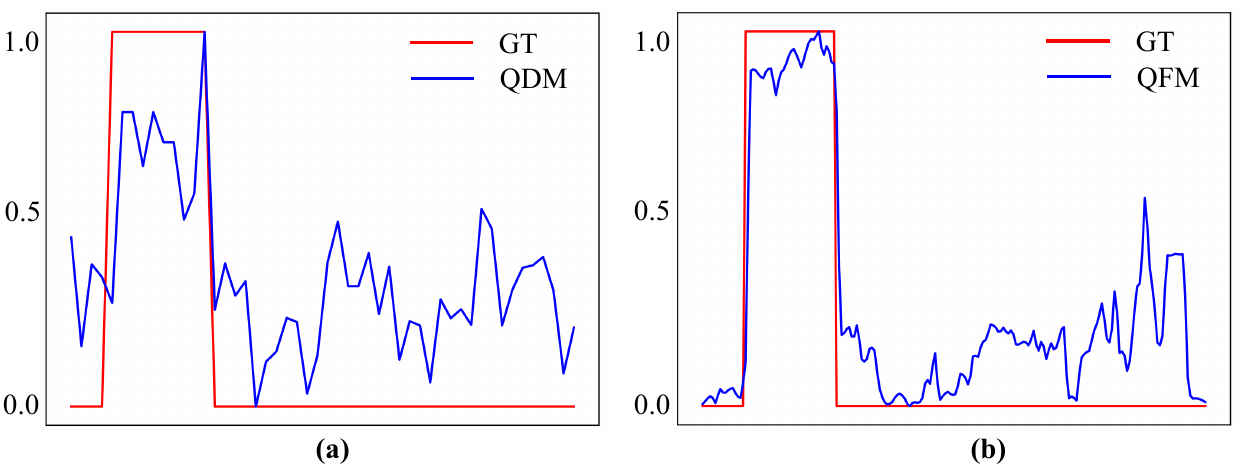}

\end{center}
   \caption{An example of the maximum and minimum normalized Query-Description Matching score (QDM) and Query-Frame Matching score (QFM) within a video of training dataset.}
\label{fig:qcm}
\end{figure}

We show an example of Fig. \ref{fig:qcm} (a) the matching score of the query corresponding to a video with the description generated by MLLM for each frame, and Fig. \ref{fig:qcm} (b) the matching score of the query with the video frame calculated by the MLLM.
It can be seen in Fig. \ref{fig:qcm}, generally the video frames corresponding to the query would generates descriptions that are highly similar to the query, and getting high query-frame matching score.

\noindent \textbf{Scalability Analysis} Our method can be easily extended to existing WSVG models and improve their grounding performance. 
We extend our method to three existing WSVG methods, including two query-based reconstruction methods, CNM \cite{zheng2022weakly} and CPL \cite{zheng2022weakly1}, and one MIL-based method, RSTPN \cite{zhang2020regularized} on ANC dataset.
Notably, since RSTPN uses a predefined manual pseudo boundaries, the PCL loss proposed cannot be applied, and we only generalize the Boundary Expand and Mutual learning to RSTPN.
As shown in Tab. \ref{tab:Scal} (a), we can
clearly conclude that the proposed framework
can greatly improve the performance of three existing methods, verifying the effectiveness our method.

\noindent \textbf{Effectiveness of EtC framework}
To further explore the effectiveness of EtC, we compare various pseudo-label generation methods including our MLLM, in the EtC pipeline. 
The iterative pseudo-label strategy (IPL) uses the training results from the previous epoch as pseudo-labels for the current epoch. However, this approach is susceptible to the impact of error accumulation.
The Mean Teacher framework \cite{tarvainen2017mean} (MT) employs data augmentation strategies, updating the teacher network with exponential moving averages to supervise the student network. 
Besides, UGS \cite{huang2023weakly} designs uncertainty based mean teacher framework.
As shown in Tab. \ref{tab:comp} (b), our proposed method exhibits the most significant performance improvement. 
Using other pseudo-label methods can also outperform the baseline, proving the effect of EtC.
We also find that using MLLM achieves better performance than other pseudo-label generation methods, showing the superiority of MLLM.
That's beacuse the use of MLLM generates precise descriptions of frames within pseudo boundaries region, influencing the grounding model to produce accurate and comprehensive pseudo boundaries, thereby enhancing model performance. 
Additionally, the PCL loss plays a vital role in constraining the pseudo boundaries's position, effectively addressing error accumulation from the initial low-quality pseudo boundaries.
\begin{table}[!tbp]
    \centering
    \caption{The impact of PCL on other pseudo-label method on C-STA dataset. `AVG IoU' represents the average IoU ratio between output proposals and GT. `$\Delta$' refers to the R1@IoU 0.5 performance improvement.}

     \scalebox{0.9}[1.1]{
    \begin{tabular}{c|c|c|c}
        \hline
{method}&R1@IoU0.5(\%) &AVG IoU(\%)& $\Delta$ (\%) \\ \hline
        Base / {(w. PCL)} &49.24 / 50.73	&41.42& {+1.49} \\ \hline
        Base + Mean-Teacher / {(w. PCL)} 	& 50.68 / 51.36&42.33	&{+0.68}\\  \hline
        Base + MLLM / {(w. PCL)} 	&53.07 / 53.39&46.81  & {+0.32}\\
        \hline
        \end{tabular}
        }
         \label{tab:pcl1}
         \vspace{-0.7cm}
\end{table}
\noindent \textbf{Ablations of the MLLM and Text Information Expand.} We conduct analytical experiments of the MLLM and Text Information Expand on ANC dataset, using CPL\cite{zheng2022weakly1} as the baseline. 
We discuss the impact of the number of generated descriptions $n_{p}$ per frame during the auxiliary text description generation stage.
The result in Tab. \ref{tab:np} (a) shows that the best performance is achieved at $n_{p}=5$.
Additionally, to obtain the description for the proposed region, we usually randomly sample $n_{f}$ frames from within the pseudo boundaries region. 
From these $n_{f} * n_{p}$ descriptions, we randomly sample one to serve as the description for that region.
The results in Tab. \ref{tab:nf} (b) show that the performance the model performs optimally when $n_{f}=5$.
Besides, the descriptions generated by different MLLM also impact the model results. As shown in Tab. \ref{tab:mllm} (c), the descriptions produced by BLIP2 result in superior performance.
\begin{table*}[!htbp]
    \centering
    \caption{Hyper-parameter sensitivity analysis on ANC dataset, where the CPL as the baseline.}

    \begin{minipage}{0.45\linewidth}
	\centering
 \scalebox{1.2}[1.2]{
        \begin{tabular}{c|ccc}
        \hline
        $\alpha$&R1@0.1&R1@0.3&R1@0.5\\ \hline
        0 & 81.03 & 57.70 & 34.16\\
        0.25&83.93 &\textbf{61.31} & \textbf{37.01}\\
        0.5 &83.38 & 61.26 & 34.63\\
        0.75& \textbf{84.16} & 61.00 & 35.09\\
        1 &82.19 & 60.72 & 34.91\\
        \hline
        \end{tabular}
        }
        \caption*{(a) The impact of parameter $\alpha$ on the final grounding results.}
         \label{tab:qfm1}
        \end{minipage}
        \hfill
	\begin{minipage}{0.45\linewidth}
		\centering
   \scalebox{1.2}[1.2]{
        \begin{tabular}{c|ccc}
        \hline
        $\beta$&R1@0.1&R1@0.3&R1@0.5\\ \hline
        0 & 83.07&60.87 &35.05\\ 
        0.025& 83.51 & 60.37 & 35.59 \\
        0.05 &\textbf{83.93} & 61.31 & \textbf{37.01}\\
        0.075& 83.76 & 61.52 & 34.44 \\
        0.1& 82.80 & 61.07 & 34.92 \\
        \hline
        \end{tabular}
        }
        \caption*{(b) The impact of parameter $\beta$ on the final grounding results.}
        \label{tab:spt}
	\end{minipage}
    \vspace{-1cm}
\end{table*}

\noindent \textbf{Ablations on the PCL.} 
We conduct ablation studies of PCL on ANC dataset, using CPL \cite{zheng2022weakly1} as the baseline.
We compare the performance influence of Query-frame (QFM) and Query-description (QDM) matching score in PCL.
As shown in Tab. \ref{tab:qfm} (a), both QFM and QDM matching score sequence can effectively aiding the model in refining the learnable pseudo boundaries, and the model achieves optimal performance when used simultaneously.
Besides, we also analyzed the impact of hyper-parameters $\tau$ and the threshold $\delta$ on the PCL loss.
As shown in Tab. \ref{tab:t} (b) and (c), the best performance is achieved at $\tau=0.25$ and $\delta=0.15$, individually.
Additionally, PCL is designed to improve low-quality pseudo-boundaries. The better the initial quality of the pseudo boundaries (i.e., closer to the ground truth), the less impact PCL has, and the opposite is true for lower-quality boundaries. 
To demonstrate PCL's effectiveness, we also compared it with other pseudo label methods in Tab. \ref{tab:pcl1}. Our findings show that PCL significantly enhances lower-quality pseudo boundaries, as seen in our results table. For instance, for initially low-quality pseudo boundaries, PCL managed to increase their quality by 1.49\% and 0.68\% for Baseline and Mean-Teacher methods, respectively.

\noindent \textbf{Analysis of Reliable results.} In Fig. \ref{fig:GT}, we demonstrate the sample count at different ratios of the intersection length between pseudo boundaries and GT to the length of GT, using CPL \cite{zheng2022weakly1} as the baseline model.
Although the initial pseudo-boundaries indeed located near significant frames within the GT boundaries, they are not complete enough.
The intersection length between the baseline model's output pseudo boundaries and the ground truth (GT) is often shorter than the length of GT. 
In contrast, the proposed framework significantly increases the intersection length between the model's output pseudo boundaries and GT. 
The result of Fig. \ref{fig:GT} shows our method has significantly reduced unreliable pseudo-boundaries compared to the baseline, which proves the our method's effectiveness in mitigating the impact of inaccurate pseudo-labels on the model. 
Specifically, the percentage of unreliable pseudo-boundaries (Intersection with GT is 0) decreased from 23.55\% in baseline to 17.89\% in ours on C-STA, and from 12.77\% to 6.32\% on ANC.
This indicates that adopting the proposed framework can effectively alleviate some of the grounding issues.

\noindent \textbf{Hyper-parameter sensitivity analysis.} In addition to the above, this paper also designs two hyper-parameters $\alpha$ and $\beta$ to balance the final loss function.
We tested the sensitivity of these hyper-parametric design ablation experiments respectively.
As shown in Tab. \ref{tab:spt}, the model achieve best performance on $\alpha=0.25$, $\beta=0.05$.

\subsection{Qualitative results}
\begin{figure}[!htbp]
\begin{center}
    \includegraphics[width=1.0\linewidth]{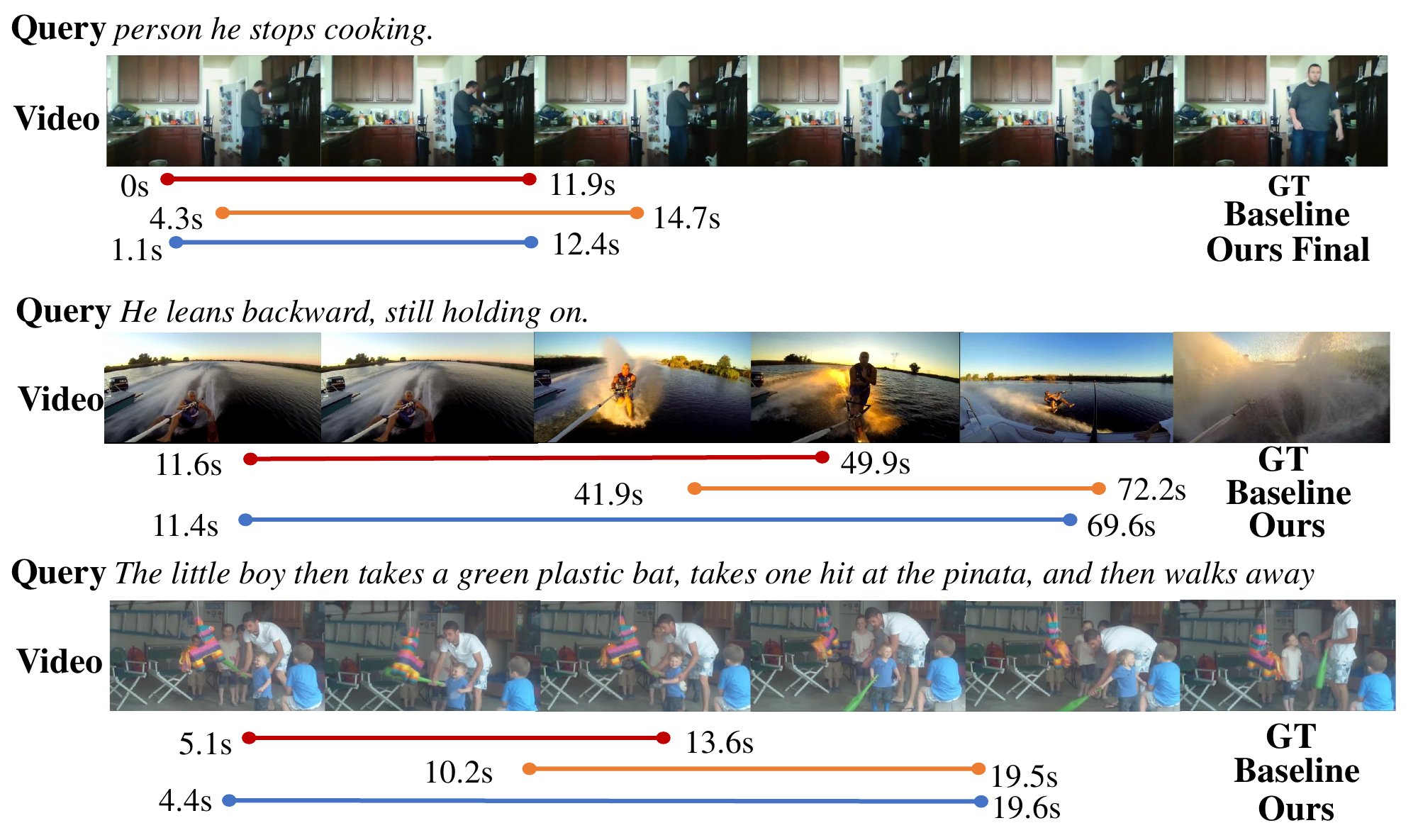}
    \vspace{-0.1cm}
\end{center}
   \caption{Qualitative results of the ground truth (GT), the Baseline model, and the Baseline model with our EtC framework. The first examples are from the Charades-STA dataset, and the last two examples are from the ActivityNet Caption dataset.}
\label{fig3}
\vspace{-0.1cm}
\end{figure}

\begin{figure}[!htbp]
\begin{center}
    \includegraphics[width=1.0\linewidth]{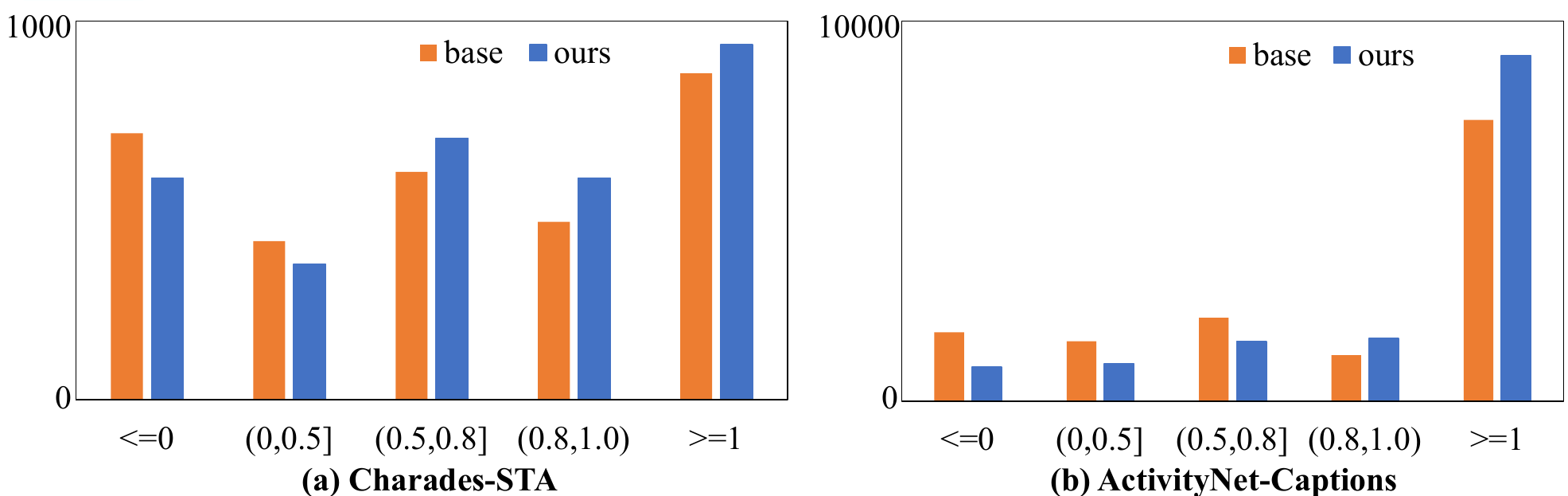}
    \vspace{-0.1cm}
\end{center}
   \caption{The sample count at different ratios of the intersection length between pseudo boundaries and GT to the length of GT. Horizontal coordinate denotes the ratios of the intersection length between pseudo boundaries and GT to the length of GT. Vertical coordinate denotes the number of test samples. The baseline model is CPL.}
\label{fig:GT}
\vspace{-0.1cm}
\end{figure}

As shown in Fig. \ref{fig3}, compared with the original grounding model, the Completeness of the grounding model that coincides with the ground truth has increased significantly after using our EtC framework, which indicates that the proposed framework effectively alleviates partial grounding.


\section{Conclusion}
In this study, we present a novel method that maintains the integrity of original temporal content while incorporating additional information to extend incomplete boundaries.
We utilize advanced multi-modal large language models (MLLMs) for frame annotation within initial pseudo boundaries, resulting in detailed expansions.
To address noise in these expansions, we combine mutual learning with a tailored PCL approach, achieving a balance between initial and expanded boundaries for enhanced precision. 
Besides, using MLLMs indeed demands more resources and time, it provides a more detailed understanding for videos, resulting in substantial enhancement for pseudo labels.
Our comprehensive experiments on WSVG datasets validate our method's efficacy.
On the other hand, our framework refines the most confident pseudo boundary generated by a WSVG model, hence it does not improve the rank-5 recall rate for multiple boundaries.

\ifCLASSOPTIONcaptionsoff
  \newpage
\fi



\bibliographystyle{IEEEtran}
\bibliography{bibtex/bib/IEEEexample}
%

% that is all folks
\end{document}